\documentclass[conference,a4paper]{IEEEtran}
\IEEEoverridecommandlockouts

\ifCLASSINFOpdf
  \usepackage[pdftex]{graphicx}
\else
\fi
\usepackage{url}
\usepackage{balance}
\usepackage{okumoto}
\renewcommand\IEEEkeywordsname{Keywords}

\begin{document}

\title{Realizing Half-Diminished Reality from Video Stream of Manipulating Objects}

\author{\IEEEauthorblockN{Hayato Okumoto, Mitsuo Yoshida and Kyoji Umemura}
\IEEEauthorblockA{Department of Computer Science and Engineering\\
Toyohashi University of Technology \\
Toyohashi, Aichi, Japan\\
h153317@edu.tut.ac.jp, yoshida@cs.tut.ac.jp, umemura@tut.jp
}}

\maketitle
\IEEEpubid{\makebox[\columnwidth]{978--1--5090--1636--5/16/\$31.00~\copyright~2016 IEEE \hfill }\hspace{\columnsep}\makebox[\columnwidth]{\hfill }}

\begin{abstract}
When we watch a video, in which human hands manipulate objects, these hands may
obscure some parts of those objects. We are willing to make clear how the objects
are manipulated by making the image of hands semi-transparent, and showing the
complete images of the hands and the object. By carefully choosing a Half-Diminished
Reality method, this paper proposes a method that can process the video in real
time and verifies that the proposed method works well.
\end{abstract}

\begin{IEEEkeywords}
Video Enhancement; Virtual Reality; Diminished Reality
\end{IEEEkeywords}

\section{Introduction}
When we watch how objects are manipulated using surveillance cameras, hands
usually conceal or obscure some parts of those manipulated objects. Realizing
Half-diminished reality in this paper means to generate a video where both the
hands and all parts of the objects are shown at the same time. An example of
this situation is a broadcast of a Japanese chess match, where players
manipulate pieces and that an entire understanding of where those pieces are
situated is always important in order to understand the current move. In spite
of this importance, at the timing of when the move is decided, the corresponding
piece is hidden by the player's hand. This situation always happens for a video
showing when objects are manipulated by hand. This problem, in the case of a
chess game, is caused by a situation where the hands (foreground image) conceal
both the piece and the board (background image). One notable characteristic of
the situation is the existing interaction between the foreground image and the
background image. For example, the piece in the hand belongs to the foreground
image where it belonged to background image previously but will soon belong to
the background image once again. The timing of this transition is also important
information of the interaction.
\IEEEpubidadjcol

In this paper, we will propose an image processing method that satisfies the
following conditions. (A) The background image has no concealed areas. (B) Both the background image and foreground image are integrated into a single image. (C)
The movements of the background image and foreground image are synchronized. (D)
The processing should be in real time so that it can be used for live broadcast.
(E) The input of processing is a single video from a fixed camera.

This kind of image processing is known as Diminished Reality (DR), which is a
part of the Virtual Reality (VR) research field. DR is to show us a virtual
space, where some objects in the real world have either vanished or are
transparent. Though many researches in DR \cite{Herling2012, Lepetit2000, Zokai2003} make objects completely
transparent, some researches \cite{Buchmann2005, Sugimoto2014} focus on semi-transparency, which is called
Half-Diminished Reality (H-DR). DR focuses on obtaining a clear and sound image
of the background; whereas H-DR regards both the foreground and the background
images as important. Buchmann et al. \cite{Buchmann2005} proposed to show a H-DR image through a
head mounted display to improve the efficiency of manual operation, and
discussed the level of transparency and the efficiency of the operation.

He and Zhang \cite{He2007} proposed a notable DR method which satisfies condition D and E.
He and Zhang obtained the complete contents of a whiteboard (Background image) in a
video conference system by using a single camera. It obtained the background
image by substituting the background image with only the stable part of the
current image. Since hands (foreground image) are not usually stable, only the
written contents on the whiteboard appear in the output image. If we can obtain
a complete background image, we can obtain the H-DR image that satisfies
conditions A and B (but not C) by superimposing the background image and the
current image. Since it requires considerable time to judge whether a part is
stable or not, the update of the background image is delayed from the change of
the current image. As a result, a direct application of He and Zhang's proposal
cannot satisfy condition C. Condition C is very challenging since at the timing
of transition of the object image from the foreground image to the background
image, the hands are usually holding the objects and a true background image may
not be available at this time.
\IEEEpubidadjcol

In this paper, we focus our attention to the application of broadcasting. With
broadcasting, though it requires processing the image at the same rate of the
current image, it is acceptable if there is a delay of several seconds from
input to output. This presents an idea where we can adjust the timing of the
current image so that it can be synchronized with the background image.

The contribution of these works is as follows. (1) We define the situation of
H-DR where interactions between the foreground and background image exist. (2)
By distinguishing the delay from real time, we show that a time shift approach
is usable for real time and a synchronized setting. (3) We explain a valid
situation where the proposal is effective.

\section{Related Work}
DR processing usually consists of two steps. The first step is to distinguish
the foreground image from the background image. The second step is to complement
that part of the background image where the foreground object conceals or
obscures the background image.

\subsection{Detecting Foreground Image}
One of the well-used methods to detect the foreground image is background
subtraction. The simplest method of background subtraction method is as follows.
First, we capture the background image without a foreground object. Then, we
compare the color and brightness of the current image and the captured background.
If the difference is greater than some set threshold, then we can regard the area
as a part of foreground image. Wren et al. \cite{Wren1997} enhance this method. Instead
of using one static background image, it uses the average of several images
as the background. This method obtains the foreground object in real time,
but the result is particularly sensitive to shadows and changes in lighting.
Instead, using a single background image, the statistical model for a
background image is commonly used \cite{Piccardi2004}.
Some works use Gaussian-Mixture \cite{Haines2014, Stauffer1999, Zivkovic2004},
and other works use a non-parametric estimation using kernel \cite{Elgammal2000, Han2004, Liu2007}.
There is also a work that uses Eigen vectors from a series of images to model the background \cite{Oliver2000}.
However, all of these works have a problem when
the foreground and background are interacting. There is another work that does
not construct the color model, but uses an RGB-Depth camera in order to detect
the foreground \cite{Sugimoto2014}.
This approach requires a special camera, and is not suitable
for the application of Japanese chess where the foreground object and the
background object may be the same position.

He and Zhang \cite{He2007} use the difference between the current image and the previous
image but not the fixed background images. For a certain time, if there is
no significant difference, the part of the image is regarded as stable. If the
image is changing during a certain period, the corresponding part of the image
is treated as a foreground region. The key concept of this approach is that the
foreground image is unstable. Please note that an unstable region may not be a
foreground region, but that a stable region is always a background region.
In this approach, the background region, which is uncovered by a foreground
object, is also treated the same way as the moving foreground region.
In this sense, this approach does not correctly detect a particular
foreground region, but successfully detects the region that contains
all of foreground regions.

\subsection{Complementing Background Image}
One of the natural ways to complement a background image is to use multiple
cameras with different angles towards an object. Using the transformation of an
image, it recovers the hidden part from other videos taken from different angles~\cite{Enomoto2007, Hashimoto2010, Zokai2003}.
Apparently though, it has a problem when hands (foreground objects) hold
the objects (background objects), since there is no angle at which we can see
the objects.

Another approach is to complement the hidden background in the same context as a
neighboring background region. This approach assumes that the background is a
wall or some other uniform texture. However, this is not a case where the
background is a collection of objects, for example, a board and pieces of
Japanese chess.

He and Zhang \cite{He2007} use a time shift approach. Since only a single camera is used in
this approach, the information complementing the background image comes from the
previous images. In this approach, the region that is regarded as the foreground
part is not updated. As a result, the past images are used to complement the
region that is currently concealed by a foreground object. The background image
is updated by the current image in only the part where the image is regarded as
stable. A nice feature of He and Zhang's approach is the ability to update the
background. Please note that the change of background is inevitably delayed in
order to judge its stability. Since they aimed to recover the information from a
white board, in other words, they aimed at DR only, they did not need to
consider the interaction between the foreground image and the background image.
Therefore, this delay was not an issue for them.

\section{Proposed Method}
We aimed at an enhanced visual image from a surveillance camera showing some
hands manipulating some objects. In this case, we are willing to see the
complete shape of objects, and also see how hands manipulate objects. An object
belongs to the foreground image when it is held by hands. The same object belongs to
a background image when it is not held by hands. We are willing to show the
information about how the object is placed at the end time of the manipulation.

First, we generate a DR image, where an image of foreground objects that obscure
a background image is diminished. Then, we superimpose the current video image
onto the background image. Then we will get a H-DR image, where the foreground
objects are half-transparent. In this image, we can see both the hands and the
objects. Since some objects may move between the foreground image and the
background image, it is a requirement that both images are synchronized. Our
approach is to adjust the foreground image to obtain the final result. This
approach will work if the person who is manipulating the objects and the person who
is watching the video are independent, which is a typical case for a
surveillance camera.

When the manipulation finishes, the object is held by hand.
Therefore, a time shift approach would be a reasonable choice to complement the image of the object. We choose the He and Zhang's approach~\cite{He2007} to realize DR.

\subsection{Implementation of Diminished Reality}

We use the temporal change of color values in pixels. As is shown in \figurename\ \ref{fig_1},
when a moving foreground object covers the background image, the color value of
the corresponding image is not stable within a certain time. In this case, we
do not update the pixel by the current image. As is shown in the red line in
\figurename\ \ref{fig_1}. The foreground object is erased from output. As is shown in \figurename\ \ref{fig_2},
when an object in the background moves to another location, the first and last color values are different. While the object is moving, the color value is
changing. As before, the background image is not updated during movement and
the color value before the movement is used for output. Suppose that the
movement has finished at the time, $t$. The input color value becomes stable.
After some duration, the system estimates that the movement is finished, and
updates the output as the input value.
This estimation can be done simultaneously for the block of pixels.
We choose this block as $16 \times 16$ pixels square.
We define the length of estimating duration as
background estimation time, $d$. Please note that movement in the current image
finishes at the time $t$, while the output is updated at the time $t + d$. We need
to adjust $d$ as the method of manipulation. For the case of Japanese chess, we
choose $d$ to be 3 seconds.
This $d$ should be adjusted according to the movement of the foreground.
If the foreground is an industial robot, the $d$ could be less than 1 second.

\fig[width=0.9\columnwidth]{
A foreground object moves over the background.
Since the color value keeps changing during object movement.
If we suppress updating of the image, the foreground object is erased.
}{fig_1}{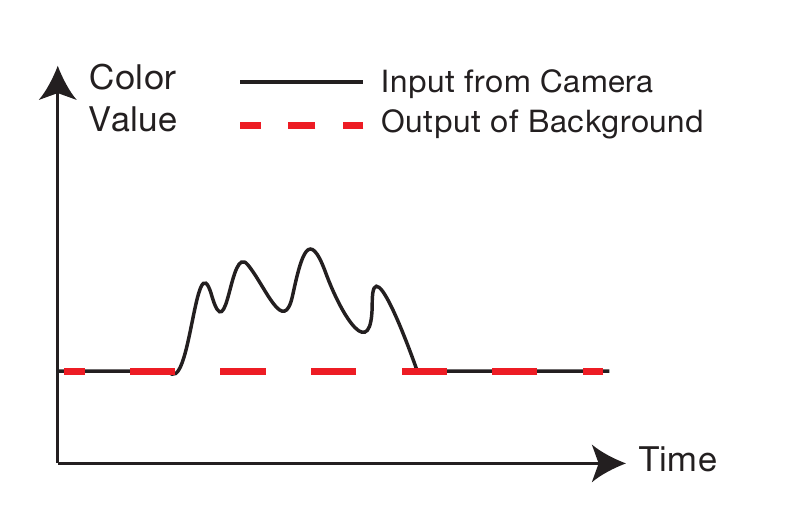}

\fig[width=0.9\columnwidth]{
An object in the background becomes a part of the
foreground, and then becomes a part of background again.
The background image will reflect the change after a certain duration passes.
}{fig_2}{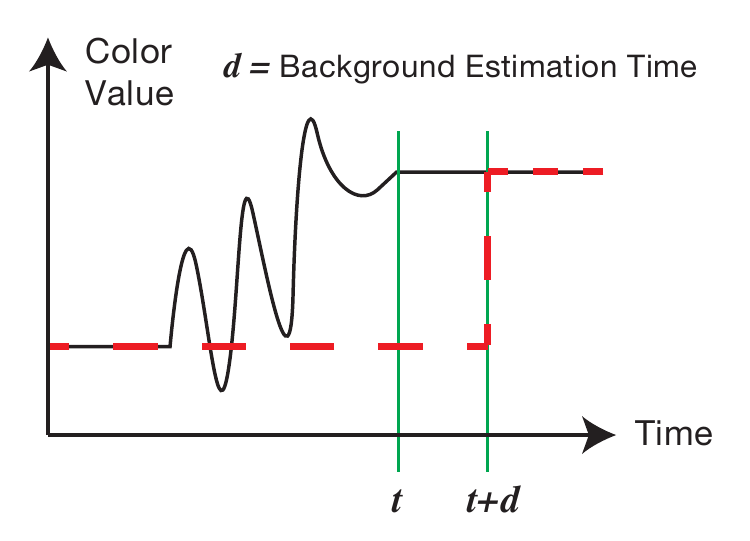}

\subsection{Implementation of Half-Diminished Reality}
Though we can see both hands and objects by superimposing the background image
and the current image, there is a problem as is shown \figurename\ \ref{fig_3}. \figurename\ \ref{fig_3} shows
the current image and the background image that are superimposed at the time $t$.
Since the update of the background image is delayed, the hand and the piece are
not synchronized. Accordingly, the output is confusing.

It is possible to delay the current image by time $d$, using a ring frame buffer.
Then we can get a delayed image as is shown as the blue line in \figurename\ \ref{fig_4}.
\figurename\ \ref{fig_5} shows the delayed image and the background image that are superimposed at the
time $t + d$. The hand and the piece are synchronized. We can see the character on
the piece at the time $t + d$, which may be informative for a video audience.

\fig[width=0.9\columnwidth]{
The result of a superimposed image of the current image and the background image at the end time of a movement. The hand and the piece are not synchronized.
}{fig_3}{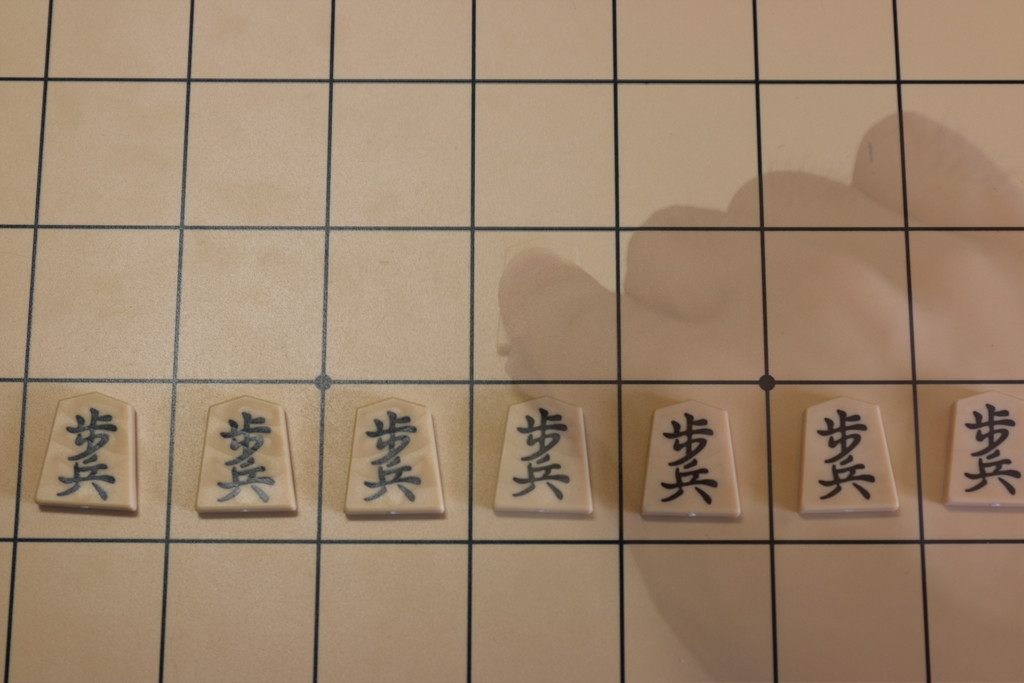}

\fig[width=0.9\columnwidth]{
Relationship between the delayed image and the background image. At the end of the movement in the delayed image, the background image changes.
}{fig_4}{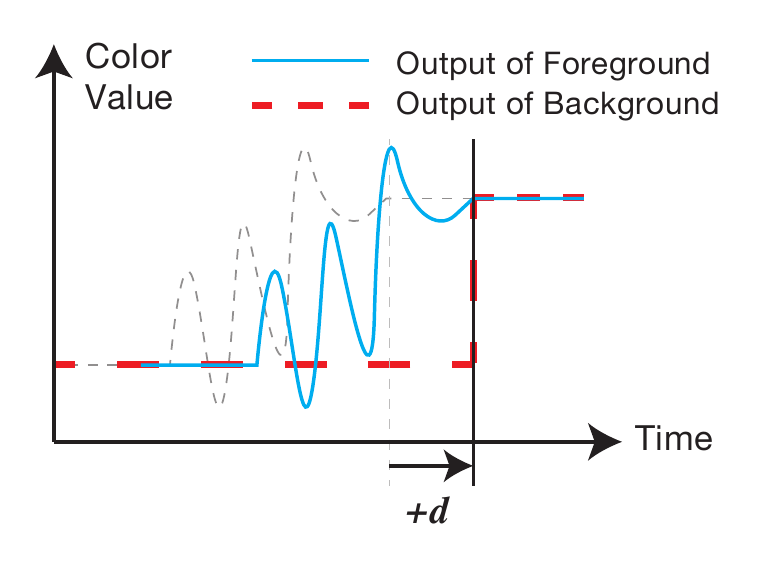}

\fig[width=0.9\columnwidth]{
The result of a superimposed image of the delayed image and the background image at the time of reflecting change. The hand and piece are synchronized.
}{fig_5}{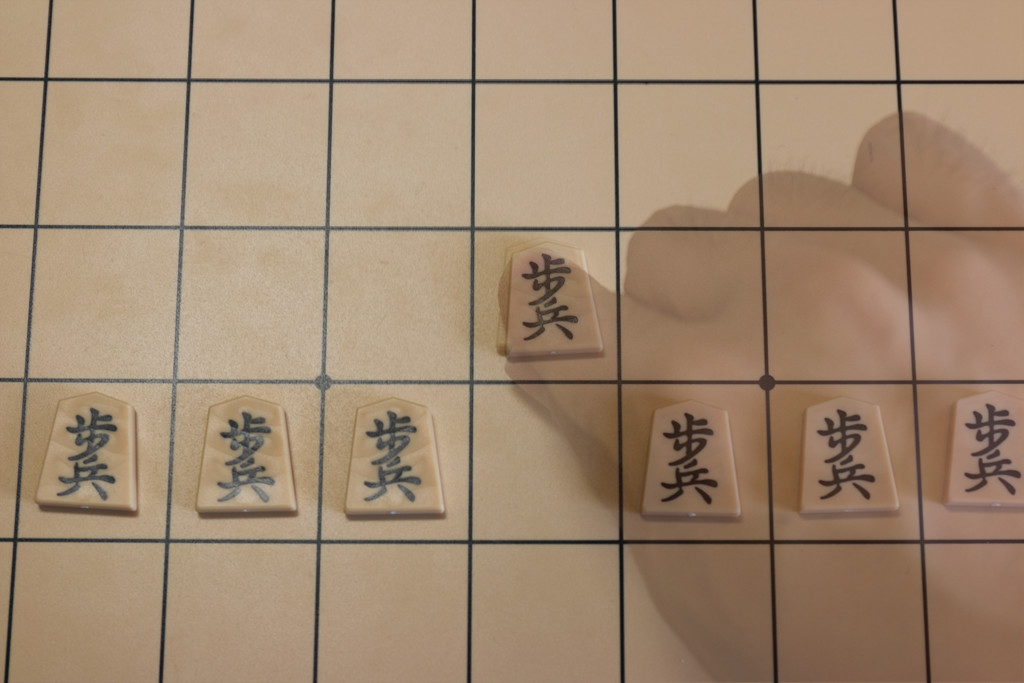}

\section{Verification of the Requirement}
\begin{figure*}[t]
  \centerline{\includegraphics[width=170mm]{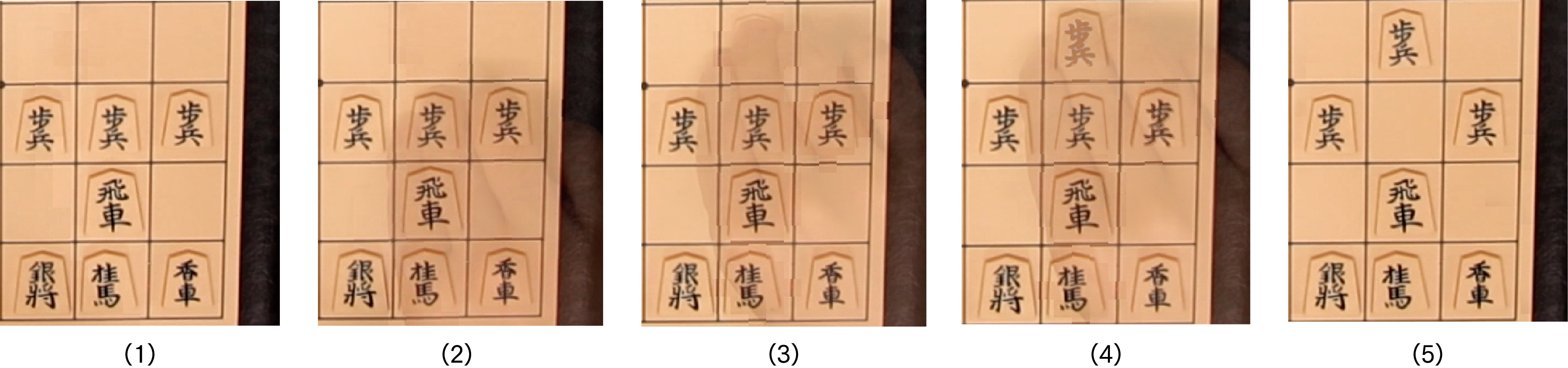}}
  \caption{
  A sequence of output video is shown.
  We can always see the characters on the pieces, and the movement of hands.
  The pieces change and the hand movements are synchronized.
  }
  \label{fig_6}
\end{figure*}

We will verify that the proposed method satisfies the conditions that we set at
the beginning. \figurename\ \ref{fig_6} shows the output sequence of  our system.

\figurename\ \ref{fig_6} (1) shows the initial situation where the hand is not in the figure.
\figurename\ \ref{fig_6} (2) shows the situation when the hand covers the board and starts the
movement. \figurename\ \ref{fig_6} (3) shows the situation in the middle of the movement, where
the piece is shown in its initial position. \figurename\ \ref{fig_6} (4) shows the situation at
the end of movement. In this figure, the piece in its initial position is about
to disappear, and the piece at its next position is appearing. \figurename\ \ref{fig_6} (5)
shows the situation where the hand is not in the figure after movement. Since we
can always see the character on the piece, condition A is satisfied. Since we
can also see the hands and piece by the superimposed image, condition B is
satisfied. As in  \figurename\ \ref{fig_6}~(4), the movement of the hand and piece is
synchronized and condition C is satisfied.

\figurename\ \ref{fig_7} and \figurename\ \ref{fig_8} show both the input and output of our system. \figurename\ \ref{fig_7}
shows the situation where the hand is about to move a piece. \figurename\ \ref{fig_8} is 3
seconds later, where a hand appears in the display. The input is captured by a
web camera. We use a 3.3GHz Intel Core i3-3220 (2 cores CPU). The resolution is
HD(1280 by 720 pixels). The output goes to network using Motion JPEG. The load
is 26\% at the frame rate 9.5 fps, using a single thread, and the load is 47\%
at a frame rate of 13 fps by multithread implementation. This implies that
conditions D and F are satisfied.

\fig[width=0.9\columnwidth]{
Both current image and output (in the display) is shown at the time when the hand starts moving the piece.
The hand is not yet appearing in the display.
}{fig_7}{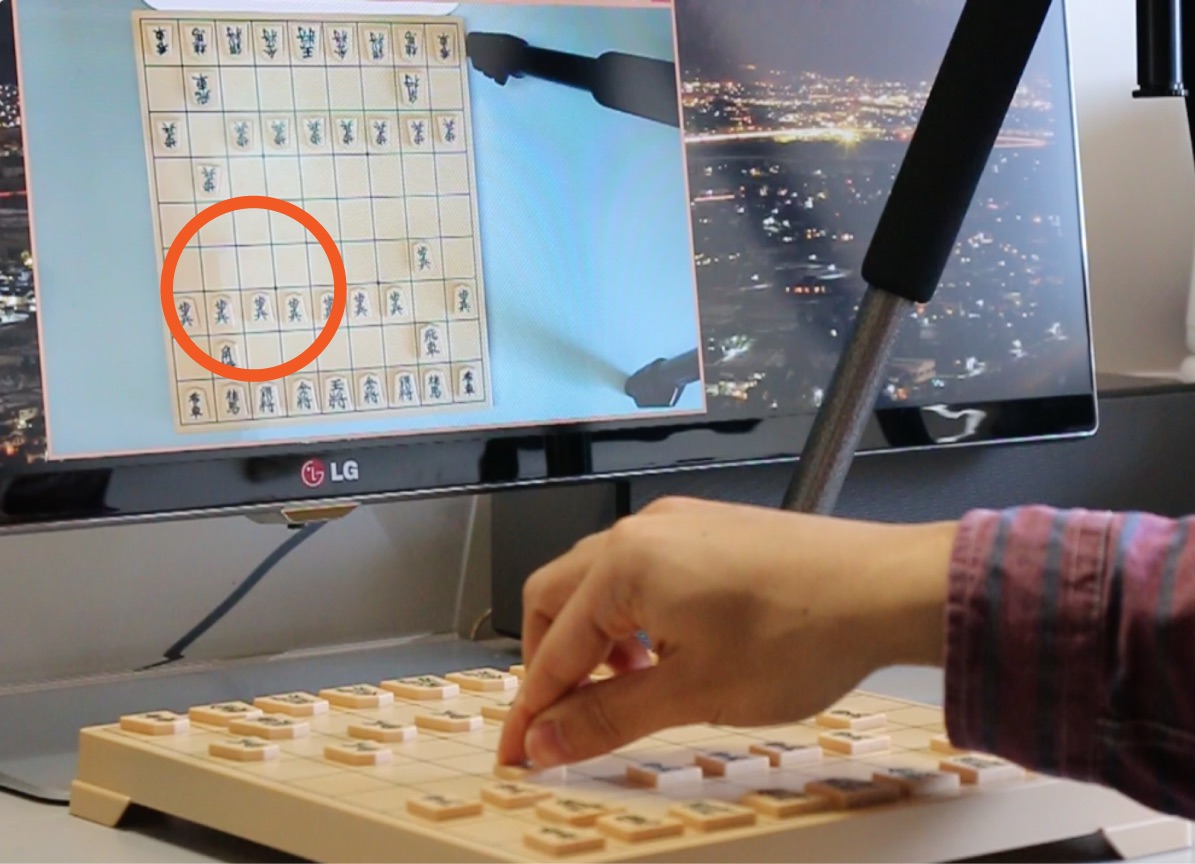}

\fig[width=0.9\columnwidth]{
After 3 seconds from \figurename\ \ref{fig_7}. The hand appears in the display.
It shows that the output is delayed but keeping up with the input.
}{fig_8}{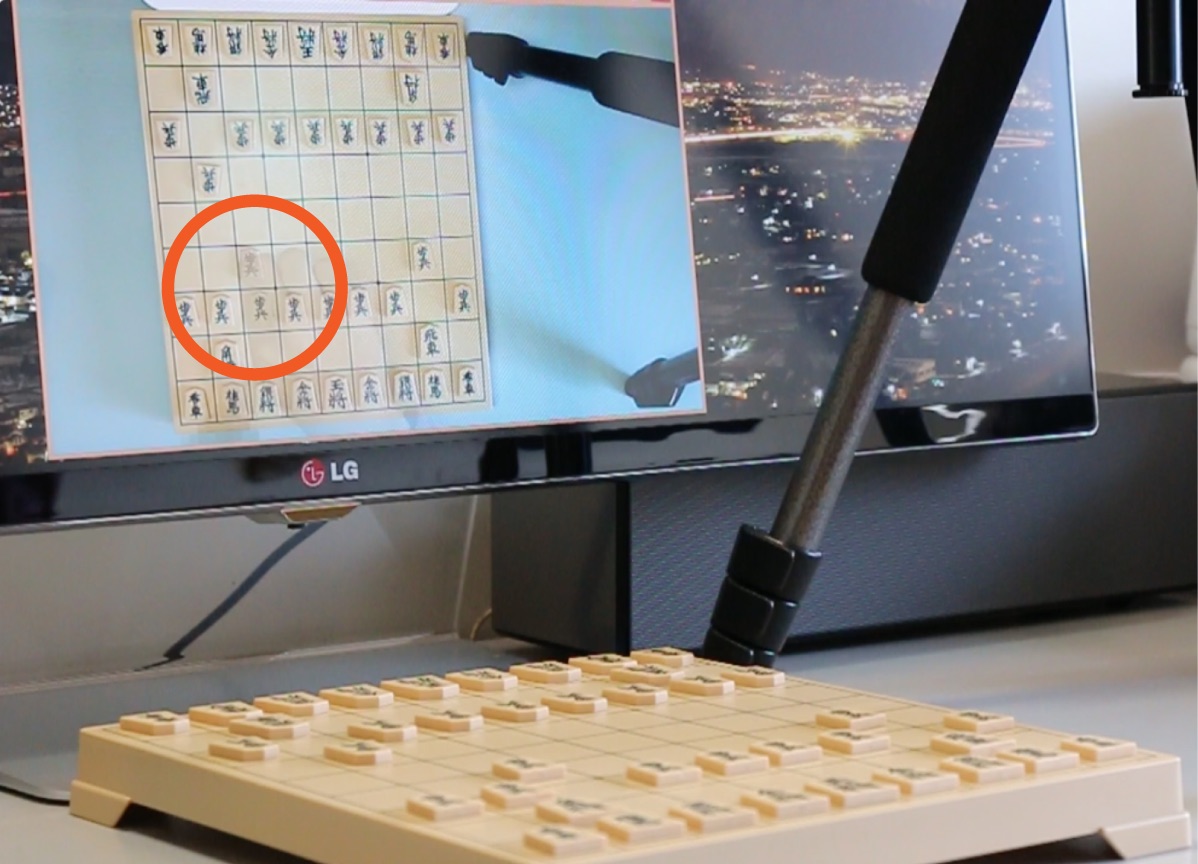}

\section{Discussions}
By this proposed method, the output is delayed from the input by several
seconds. This may limit the situation where the proposed method may be useful.
We assume that the person who manipulates the object is different from the person
who watches the output. If this assumption is not applicable, the delay
might be a problem. Nevertheless, we can assume this in many cases especially
when using a surveillance camera.

The other assumption that we have here is that the background object is always
stable and the foreground object is not stable longer than duration $d$. There may
be some cases where an object in the background moves spontaneously.
Nevertheless, we can assume this for the Japanese chess situation. It is also
the case for many assembling video.

\section{Conclusion}
In this paper, we have shown the implementation of H-DR, where there are
interactions between a background and a foreground. We propose to use movement
to realize DR, and then to adjust the current image by superimposing onto video.
It may be valuable that we are able to see the complete object image at the end
time of its manipulation. We have chosen the broadcast of a Japanese chess match
as an example. The same technique will be applicable in many situations where a
surveillance camera is used.
\balance

\section*{Acknowledgment}
A part of this research was supported by JSPS KAKENHI Grant Number 26330396.

\bibliographystyle{IEEEtran}

\bibliography{icaicta2016}

\end{document}